\documentclass[10pt,twocolumn,letterpaper]{article}

\usepackage{iccv}
\usepackage{times}
\usepackage{epsfig}
\usepackage{graphicx}
\usepackage{amsmath}
\usepackage{amssymb}

\usepackage{tikz}
\usepackage{comment}
\usepackage{amsmath,amssymb} 
\usepackage{color}
\usepackage{graphicx}
\usepackage{amsmath}
\usepackage{amssymb}
\usepackage{booktabs}

\usepackage{subcaption}

\usepackage{times}
\usepackage{epsfig}
\usepackage{adjustbox}
\usepackage{algorithm}
\usepackage{multirow}
\usepackage{algorithmicx} 
\usepackage{multirow}
\usepackage{algpseudocode}
\usepackage{soul}

\usepackage{xcolor}

\usepackage{booktabs}

\newcommand{\mat}[1]{\mathbf{#1}}

\newcommand{\beq}{\begin{equation}}
\newcommand{\eeq}{\end{equation}}
\usepackage[breaklinks=true,bookmarks=false]{hyperref}

\iccvfinalcopy 

\def\iccvPaperID{****} 

\ificcvfinal\fi

\begin{document}

\title{Cross-view Action Recognition via Contrastive View-invariant Representation}

\author{Yuexi Zhang\textsuperscript{1}, Dan Luo\textsuperscript{1}, Balaji Sundareshan\textsuperscript{1},Octavia Camps\textsuperscript{1} and Mario Sznaier\textsuperscript{1}\\
\textsuperscript{1} College of Engineering, Northeastern University, 360 Huntington Ave, Boston, MA, 02215\\
\{zhang.yuex, luo.dan1,sundareshan.b\}@northeastern.edu, \{camps,msznaier\}@coe.neu.edu
}

\maketitle
\ificcvfinal\thispagestyle{empty}\fi

\begin{abstract}
   \vspace{-0.4cm}
Cross view action recognition  (CVAR) seeks  to recognize a human action when observed from a previously unseen viewpoint. This is a challenging problem since the appearance of an action changes significantly with the viewpoint.  Applications of CVAR include surveillance and monitoring of assisted living facilities where is not practical or feasible to collect large amounts of training data when adding a new camera. 
We present a simple yet  efficient CVAR framework to learn invariant features from either RGB videos,  3D skeleton data, or both.
The proposed approach outperforms the current state-of-the-art achieving similar levels of performance across input modalities:{ 99.4\% (RGB) and 99.9\% (3D skeletons),  99.4\% (RGB) and 99.9\% (3D Skeletons), 97.3\% (RGB), and 99.2\% (3D skeletons), and 84.4\% (RGB) for the N-UCLA,   NTU-RGB+D 60, NTU-RGB+D 120, and UWA3DII datasets, respectively.}
\end{abstract}

\section{Introduction}

\label{sec:intro}

Human (single) action and activity recognition from video data have a wide range of applications including  surveillance \cite{2011_oh_large}, human-computer interaction \cite{2002_duric_integrating} and virtual reality \cite{2017_sudha_approaches}. Recent developments in deep learning and the release of general-purpose large scale datasets, such as the Kinetics Human Action Video Dataset \cite{2017_carreira,carreira2018short,smaira2020short} with up to 700 classes  and ActivityNet \cite{caba2015activitynet} with 203 activity classes and untrimmed videos, have fostered a large body of research on both action and activity recognition.

\begin{figure}[h]
\centering
\includegraphics[height= 2.30in, width=1.0\linewidth]{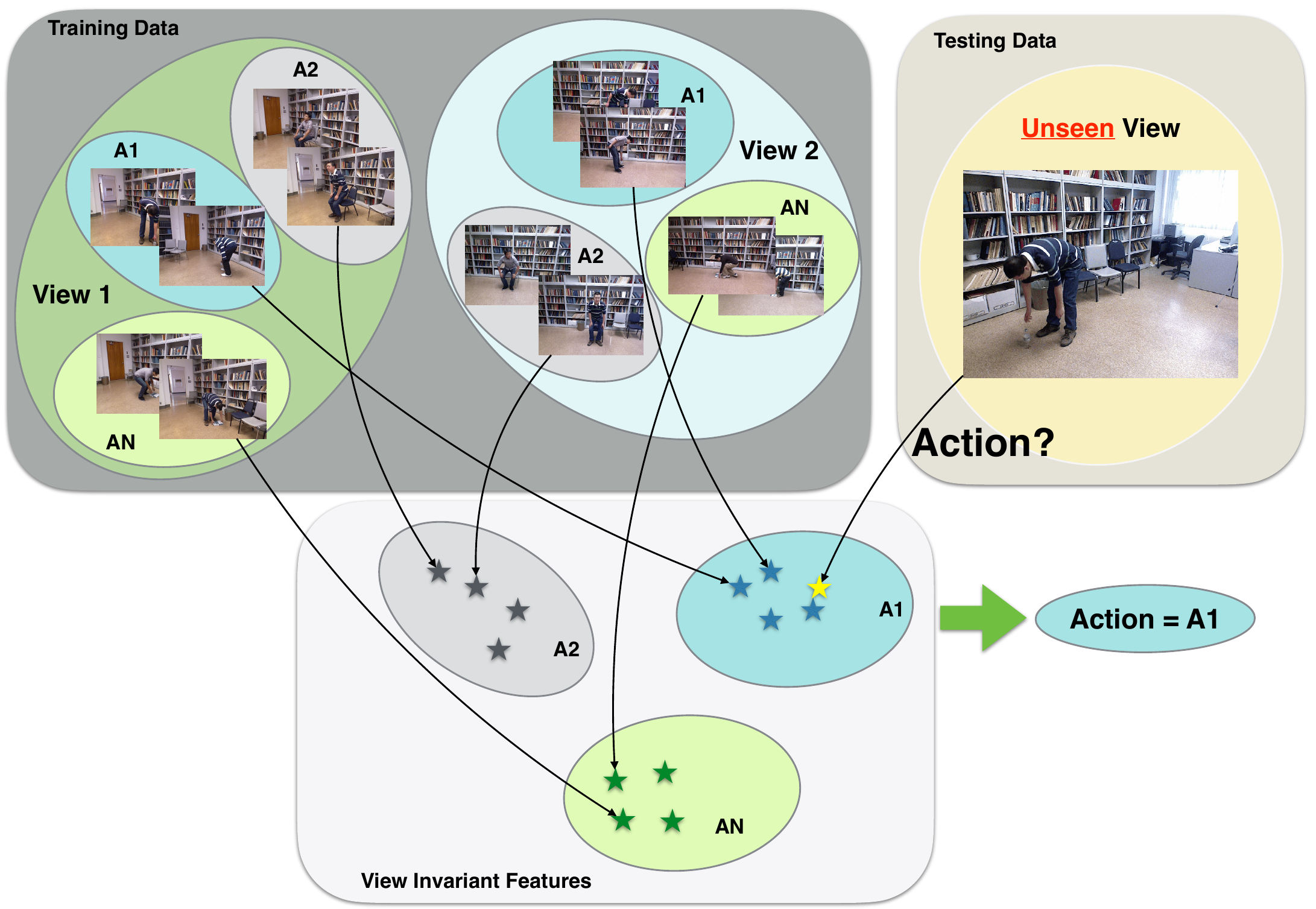}\\
\caption{\textbf{Proposed framework for Cross-view Action Recognition (CVAR).} CVAR requires making inferences using data from previously unseen viewpoints during training. The problem is challenging since actions can look significantly different from different points of view. We propose a framework where the classification is done  in a view-invariant feature space.  }
\vspace{-0.55cm}
\label{fig:CVAR}
\end{figure}

Most of the  action recognition literature \cite{huifang2021review,yao2019review} do not explicitly address the effect of view changes.  Instead, they either focus on single views,  rely on very large datasets where different viewpoints are well represented, 
or use other modalities such as 3D motion capture data or depth information which are easier to relate across views but more expensive to capture and not always available.

\begin{figure*}[t]
\centering
\includegraphics[height=3.0in,width=1.0\linewidth]{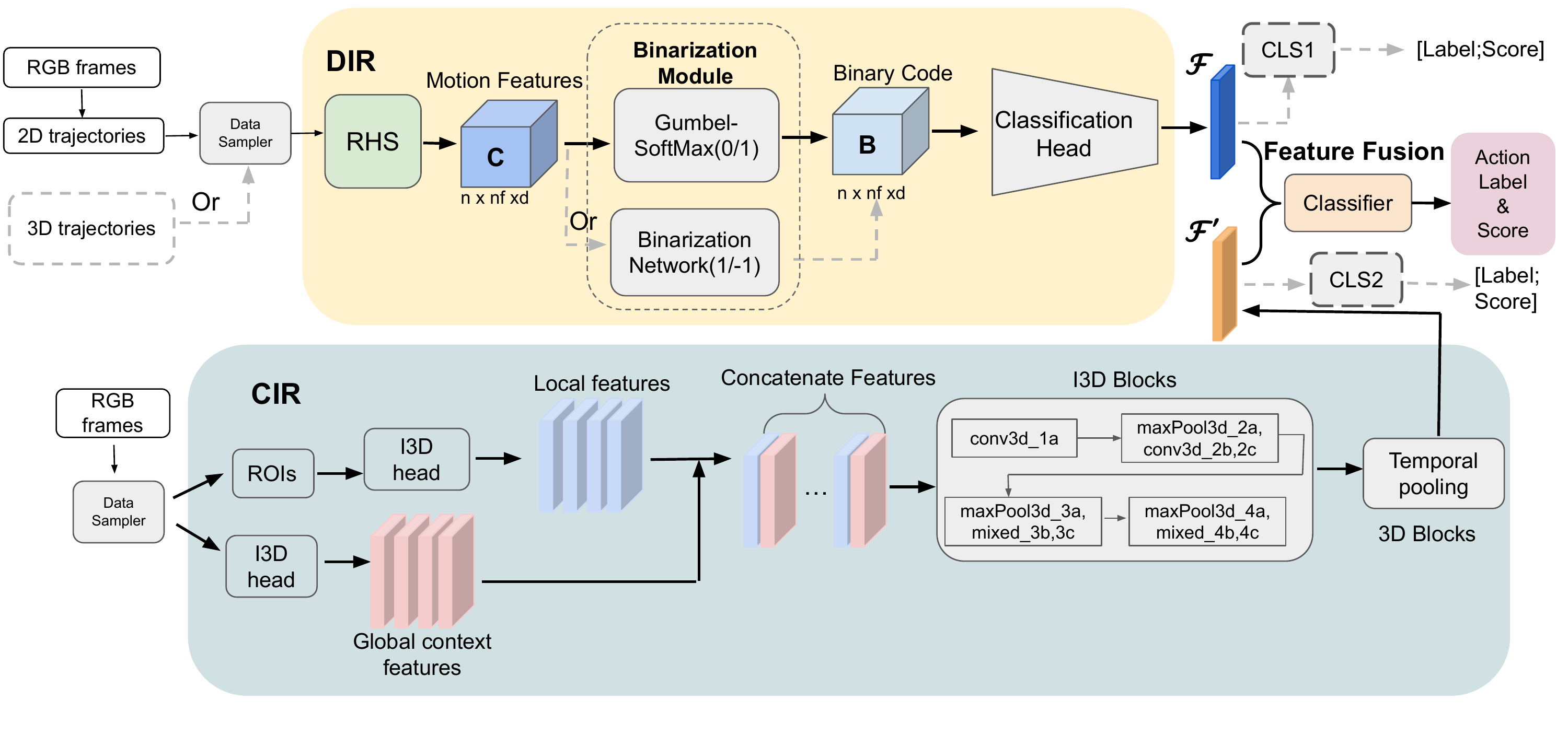}

 \caption{\textbf{Proposed architecture.} {The DIR Stream learns invariant dynamics-based features from 2D or 3D skeleton sequences. The CIR Stream learns the appearance and context of the action when RGB data is available. When applying DIR only, 'CLS1' takes features \textbf{\textit{F}} to predict probabilities for each class; when applying CIR only,  'CLS2' will return action probabilities from  \textbf{\textit{F'}}. When using the full 2-stream architecture, the action probabilities are predicted by fusing \textbf{\textit{F}} and \textbf{\textit{F'}}.}}
 \vspace{-.2cm}
\label{fig:architecture details}
\end{figure*}

In contrast, the focus of this paper is  {\em Cross-view Action Recognition} (CVAR), where the goal is to identify actions from videos captured from  {\em views entirely unseen during training}.  
CVAR is a challenging problem since the appearance of the actions can change significantly between different viewpoints,  as illustrated in Fig.~\ref{fig:CVAR}. Because of this, many approaches incorporate or rely entirely on 3D data. 
However, being able to  do CVAR using only RGB data (during training and/or inference), would open up the possibility of training with much smaller scale datasets (i.e. no need to have data from all possible views) and eliminates the need for camera synchronization and collection of expensive 3D data. Motivated by this,   we propose a novel framework (Figs.~\ref{fig:CVAR},\ref{fig:architecture details})  that captures dynamics-based information from skeleton sequences in order to perform cross view classification in a view-invariant feature space. 
The main contributions of this paper are:
\begin{itemize}
\item A flexible and lean invariance-based CVAR framework, suitable for a variety of input modalities: RGB alone, 3D skeletons alone, or a combination of both. The proposed model uses only 1.4M parameters and  11.0G FLOPS,  50\%  and 30\% less  than the previous state-of-the-art (SOTA), in the NTU-60 benchmark.
\item Our method outperforms the current SOTA  performance in four standard CVAR benchmark datasets for all input modalities and on the single-view action sub-JHMDB benchmark for RGB inputs.  Furthermore,  the level of CVAR performance is the same across all modalities, bridging a long standing performance gap between 2D and 3D based methods. 
   \item We report extensive ablation studies evaluating different design choices,  types of input data, and datasets to perform CVAR and the related problem of cross subject action recognition, where the actors in the testing data have not been seen during training. 
 \end{itemize}


\begin{figure}[t]
\centering
\includegraphics[height= 1.8in, width=\linewidth]{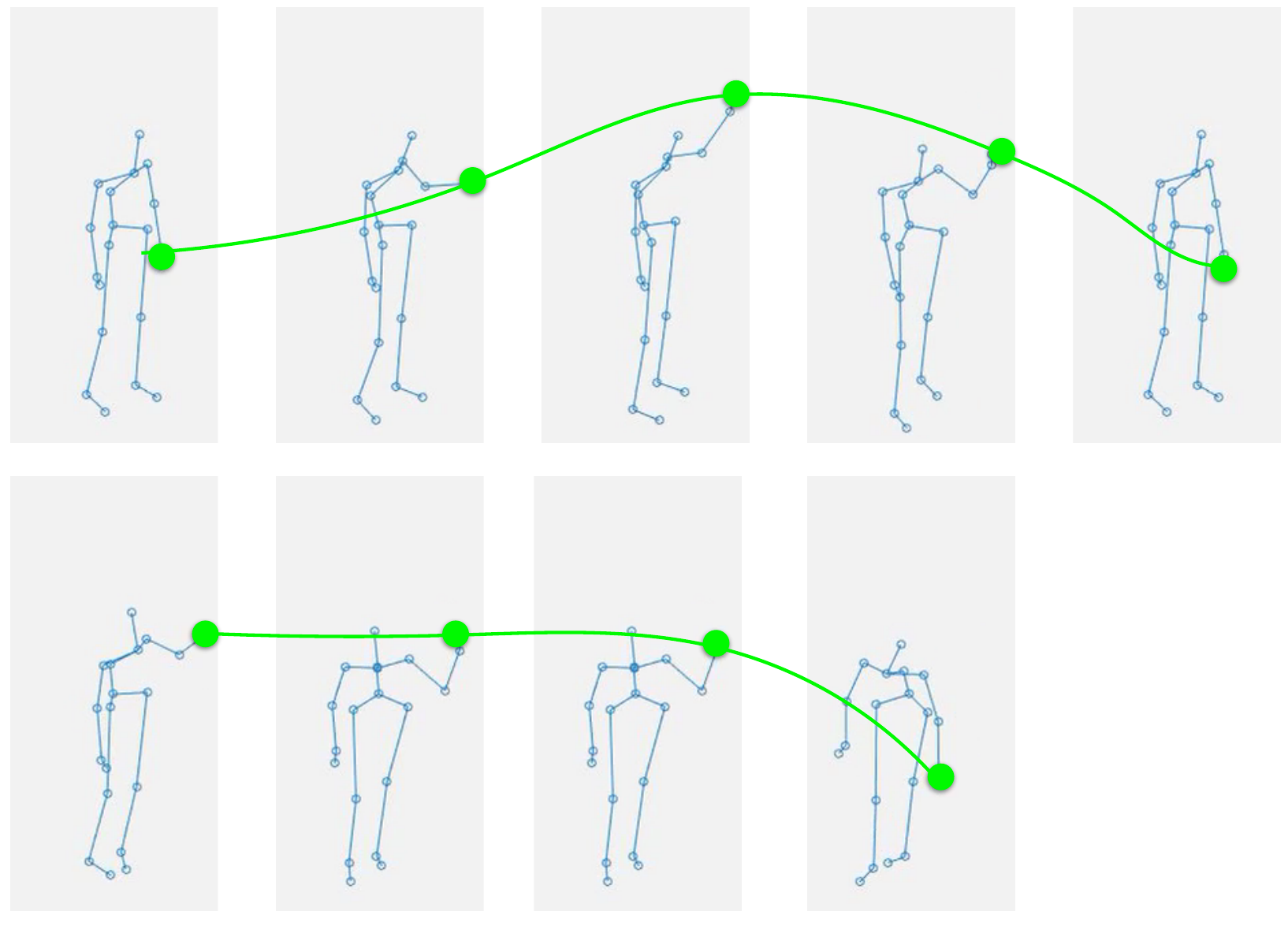}\\
\caption{\textbf{Challenges in understanding actions from skeletons:} The top and bottom frames depict two sequences (of different lengths) of the same action, observed from different viewpoints using asynchronous cameras.   It is difficult to compare  trajectories of corresponding keypoints when they have different lengths, and are  seen from different view points using unsynchronized sensors.  }
\vspace{-0.55cm}
\label{fig:Tracks}
\end{figure}

\section{Related Work}

A comprehensive review of approaches to the general topic of action recognition can be found in the recent surveys \cite{huifang2021review,yao2019review}. Here, we focus on the particular problem of CVAR, where the goal is to recognize actions from previously unseen view points. 

Many recent methods incorporate depth data or 3D skeletons, since it is easier to relate this type of information across views. Amir et al. \cite{2017_shahroudy} used a structured sparsity learning machine to explore  factorized components when RGB and Depth information are both available. Li et al. \cite{2018_li} proposed to use view-adversarial training to encourage view-invariant feature learning using only depth information. Wang et al. \cite{2017_wang_scene} extracted features from depth and RGB modalities as a joint entity through 3D scene flow to get spatio-temporal motion information. Varol et al. \cite{varol2021synthetic} proposed an approach to generate synthetic videos with action labels using 3D models. Yang et al. \cite{2021_yang_skeleton} learn skeleton representations from unlabeled skeleton sequence data using a  cloud colorization technique.  In \cite{2021_chen_channel}, Chen et al. proposed a channel-wise topology refinement graph convolution. Friji et al. \cite{2021_friji_geometric} used Kendall's shape analysis while Li et al. \cite{2021_li_else} used elastic semantic nets.  Nguyen \cite{2021_nguyen_geomnet} proposed to represent skeleton sequences using sets of SPD matrices. Su et al. \cite{2021_su_self} used motion consistency and continuity to learn self-supervised representations. 

Relatively few methods use only RGB data.  Earlier approaches used epipolar geometry to perform coarse 3D reconstruction \cite{syeda2001recognizing,roh2010view}, bag of words to get view invariant representations \cite{2011_liu_cross}, dense feature tracking to obtain view invariant Hankelet features \cite{2012_li_cross}, or used a discretization of the viewing hemisphere to learn view invariant features using shape and pose \cite{rogez2007view}. More recent approaches
use pose heatmaps \cite{2022_duan_revisiting}, codebooks  \cite{2017_kong_deeply,liu2018hierarchically,2018_rahmani_learning}, attention mechanisms \cite{baradel2018glimpse}, adversarial training \cite{marsella2021adversarial}, view-based batch normalization \cite{goyal2022cross}, 
CNN \cite{2017_ke_new,2017_liu_enhanced}, RNN \cite{2015_du_hierarchical,2017_liu_skeleton,2018_ben_coding,2019_zhang_view} and GraphCNN \cite{2018_yan_spatial,2019_li_spatio,2022_wang_skeleton} to learn view-invariant features. \cite{2010_weinland_making,2013_mahasseni_latent,2013_wu_cross} also try to achieve view invariance by using information from enough views during training and Vyas et al. \cite{2020_vyas_multiview} proposed a method using representation learning to get a holistic representation across multiple views. 
There is also a stream of approaches \cite{2012_li_discriminative,2013_zhang_cross,2013_zheng_learning,2015_rahmani_learning} that seek a view independent latent space to compare features from different views. In spite of these efforts, the performance gap between RGB-based approaches and  other modalities-based approaches remains large.

\section{Proposed Approach Overview}

Inspired by studies by Johansson \cite{johansson1973visual}, which suggest that it is possible to understand human motions by only paying attention to a few moving points, our approach will leverage recent advances in computer vision that have developed efficient and accurate  detectors of skeletons in 2D images.

In the CVAR setup, our goal is to identify actions from the motion of the human joints, captured from views entirely unseen during training.  However, this is a challenging task as illustrated in Fig.~\ref{fig:Tracks}. There, it can be seen that the raw trajectories of two corresponding joints can be significantly different when the action is observed for different amounts of time, from different viewpoints, and using asynchronous cameras. We address this challenge by learning viewpoint and initial condition dynamics-based invariant representations (DIR) that capture the underlying dynamics of the observed motions for the human joints, using only data from the training (source) views. The proposed  (DIR), which is described in section~\ref{sec:DIR}, can be used with sequences of different lengths, from either 2D or 3D trajectories, or both.

While it is true that motion alone provides strong cues for action recognition, scene context also carries useful information. Thus,  if RGB data is also available, we propose to use a two-stream approach where one branch captures the DIR from skeleton data and the other branch  captures the context information representation (CIR) from the RGB frames. The details of the CIR branch are given in section~\ref{sec:CIR}. A diagram of the complete architecture is shown in Fig.~\ref{fig:architecture details}, and its implementation details are provided in section~\ref{sec:implementation}.

\section{Dynamics-based Invariant Representation }\label{sec:DIR}

Consider two trajectories of the same human joint while performing the same action, but observed unsynchronously from different view points (as illustrated in Fig.~\ref{fig:Tracks}):
\begin{align}
\mathbf{y}^{(1)}_{1:T_1} &= [\mathbf{y}^{(1)}_1,\mathbf{y}^{(1)}_2,\dots,\mathbf{y}^{(1)}_{T_1}]^t \label{eq:y1}\\
\mathbf{y}^{(2)}_{1:T_2} &= [\mathbf{y}^{(2)}_1,\mathbf{y}^{(2)}_2,\dots,\mathbf{y}^{(2)}_{T_2}]^t
\label{eq:y2}
\end{align}
where $\mathbf{y}^{(i)}_k = (x^{(i)}_k,y^{(i)}_k,z^{(i)}_k,1)^t$ or  $\mathbf{y}^{(i)}_k = (x^{(i)}_k,y^{(i)}_k,1)^t$ are the 3D or 2D joint's homogeneous coordinates  of the $k^{th}$ observation from viewpoint $i$, respectively. These trajectories are observations of corresponding joints, and hence we can assume that  they are related by a linear transformation, once they are temporally aligned:
\begin{equation}
    \mathbf{y}^{(2)}_k = \mathbf{A} \mathbf{y}^{(1)}_{k + \delta} \label{eq:affine}
\end{equation}
where $\delta$ is the (unknown) temporal delay between viewpoints, and the (unknown) matrix $\mathbf{A}$ is a $4 \times 4$ rotation and translation transformation $\mathbf{A} = [ \mathbf{R} | \mathbf{T}]$ if both trajectories are 3D, a $ 3\times 4$ affine matrix, if one of them is an affine 2D projection of the other, or a $3 \times 3$ affine matrix if both trajectories are 2D affine projections of the 3D motion.

In this paper we will model each  trajectory as the impulse response of a discrete linear time invariant (LTI) system of (unknown) order $n_i$, with transfer {matrix} in the frequency domain $\mathbf{\cal{Y}}^{(i)}(z) = \frac{\mathbf{N}^{(i)}(z)}{D^{(i)}(z)}$, where $D^{(i)}(z)$ and the the entries of the vector $\mathbf{N}^{(i)}(z)$ are polynomials of degree  $n_i$.


\noindent {\bf Theorem 1:} Given  two corresponding temporal sequences (\ref{eq:y1}) and (\ref{eq:y2}) satisfying (\ref{eq:affine}), generated from observable LTI systems and such that {$T_i \geq 2n_i+1$}, in the absence of noise, then, the denominators of their transfer {matrices} are the same, i.e. $n_1 = n_2$ and  $D^{(1)}(z)= D^{(2)}(z)$.

\noindent {\bf Proof:} Please see supplemental material.

\noindent {\bf Corollary 1:} Since the denominator   of the transfer functions for both trajectories are identical,  their roots, i.e.  the poles of the corresponding systems, $p_1,p_2,\dots,p_n$ are also the same: $D^{(1)}(z) = D^{(2)}(z) = \Pi_{i=1}^n (z - p_i)$. 

\noindent{\bf Remark:} Comparing the raw sequences themselves is meaningless: they can be very different, even though they are from the same joint and they might have different lengths.  The above corollary provides a principled way of comparing them by comparing instead the poles of the underlying dynamics, since they are invariant to affine viewpoint and to initial conditions changes and are independent of the sequence length. Both types of invariances are relevant to the CVAR problem. Affine invariance provides support for a view agnostic dynamic encoding of the input data, while initial condition invariance shows that this representation is valid, even when the data from different views are not synchronized, or might be of different lengths, for example with one view showing only a portion of the action.


\subsection{Design of the DIR branch} 

The input to the DIR branch is a set of  motion sequences. For example, they can be $2M$ sequences with  the $x$ and $y$ coordinates for $M$ joints  as detected by an off-the-shelf pose estimator such as Openpose \cite{2017_cao_realtime}, or $3M$ sequences with the $x,y,z$ joint coordinates  measured by a 3D motion capture sensor over time.   
This input is processed by three main modules (top Fig.~\ref{fig:architecture details}), as described in detail next.

\noindent $\bullet$ {\bf \underline {RHS}:} The RHS module encodes the input sequences using a Re-weighted Heuristic Sparsity optimization layer to find fixed length, sparse representations of the inputs. This is the first step towards identifying the invariant poles and it is motivated by the observation that 
 the z-transform of the impulse response of each of the input sequences could be written as the sum of $n$ impulse responses, one for each of its invariant poles, if the poles were known: 
\[
\mathbf{{\cal Y}}(z) = \frac{\mathbf{N}(z)}{\Pi^n_{i=1}(z-p_i)} = \sum_{i=1}^n \frac{\mathbf{c}_iz}{z-p_i}
\]
Taking the inverse of the $z$-transform, we can write: $\mathbf{y}_k = \sum_{i=1}^n p_i^{k-1}\mathbf {c}_i$.
Collecting the equations for $k=1,\dots,T$:
\begin{equation}
\mathbf{y}_{1:T} = \left [ 
\begin{array}{cccc} 
1 & 1 & \cdots & 1 \\
p_1 & p_2 & \cdots & p_n \\
\vdots & \vdots & \cdots & \vdots \\
p_1^{T-1} & p_2^{T-1} & \cdots & p_n^{T-1}\\
\end{array}
\right ]
\left [
\begin{array}{c} \mathbf{c}^t_1 \\ \mathbf{c}^t_2 \\ \vdots \\ \mathbf{c}^t_n \end{array}
\right ] = \mathbf{P_y}\mathbf{C_y}
\label{eq:C}
\end{equation}
where the matrix $\mathbf{P_y}$ is  invariant, since it is completely determined by the invariant poles. 

However,  neither the number of poles $n$ nor the poles themselves are known a-priori. Thus, the RHS module uses an  over complete (to be learned) dictionary of candidate poles $\mathcal{D}_N = \left \{1,\rho_1,\dots,\rho_N\right \}$ with $N >> n$ to select a  subset $\mathcal{D}_n$ of up to $n$ poles to minimize the reconstruction error: 
\[
\left \{p_1^*,\dots,p_n^* \right\} = \arg\min_{\mathcal{D}_n \subset \mathcal{D}_N}\left \{\min_\mathbf{C_y}\| \mathbf{y}_{1:T} - \mathbf{P}_{\mathcal{D}_n}\mathbf{C_y}\|^2_2 \right \}
\] 
where $\mathbf{P}_{\mathcal{D}_n}$ is the matrix formed from the poles in $\mathcal{D}_n$.
Since the outer minimization is a combinatorial optimization problem (due to the need to select $n$ poles out of the possible $N$, where $n$ is not known), the RHS module jointly solves for the poles and $\mathbf{C_y}$ by  optimizing:
\[
\min_\mathbf{C_y} \|\mathbf{y}_{1:T} - \mathbf{P}_{\mathcal{D}_N}\mathbf{C_y}\|^2_2 
+\lambda\|\mathbf{C_y}\|_1
\]
where the first term of the  minimization objective penalizes the reconstruction error and the second term penalizes high order systems. Then, the  order of the system $n$ is given by the number of non-zero elements of $\mat{C_y}$, while the poles $\left \{p_1^*,\dots,p_n^* \right\}$ are those associated with the corresponding columns of $\mathbf{P}_{\mathcal{D}_N}$.
In \cite{2018_liu_dyan}, they solve a similar  problem using the  FISTA \cite{beck2009fast} algorithm. Our experiments show that in practice, using FISTA results on most of the elements of $\mathbf{C_y}$ to be small but non-zero, leading to overfitting.  We addressed this problem by  further promoting sparsity by introducing a re-weighted heuristic approach \cite{mohan2010reweighted} where we  run the FISTA optimization module repeated times instead of only once.  Each time, we increase a penalty applied to any small but non-zero coefficients from the previous iteration to push them closer to zero in the current iteration. This is easily accomplished by starting from the previous solution and using the inverse of the magnitude of the coefficient as its penalty. Moreover, since each iteration starts from the previous solution, the increased computational cost of running FISTA again is small. Finally, the   loss function to learn  $\mathcal{D}_N$ is:
\begin{equation}
\mathcal{L}_{D} = \|\mathbf{Y} - \mathbf{P}_{\mathcal{D}_N}\mathbf{C}\|^2_2 + \lambda \|\mathbf{C}\|_1
\label{Ldyan}
\end{equation}
where $\mathbf{Y}$ is a matrix with all the input joint trajectories.


\noindent $\bullet$ {\bf \underline {Binarization Module}:} Different from  \cite{2018_liu_dyan}, we are not interested on the matrix $\mathbf{C}$ since it  is not affine invariant. This is easy to see since, in general, $\mathbf{A}\mathbf{Y} = \mathbf{A}\mathbf{P} \mathbf{C} \ne \mathbf{P} \mathbf{C} = \mathbf{Y} $. Instead, here we seek the poles  selected by the non-zero elements of  $\mathbf{C}$. To this effect,  DIR uses a binarization module to find an indicator vector $\mathbf{b}$ for each input sequence of dimension $N$. Its bit $\mathbf{b}_k$ is turned ``on" if the value of $\mathbf{c}_k$ is non-zero to indicate  that pole $\rho_k \in \mathcal{D}_N$ is needed, and turned ``off" otherwise. Note that an added benefit of using this representation is that while the order of the underlying system  and number of selected poles $n$ can change from sequence to sequence, the dimension of the indicator vector is fixed and set to the size $N$ of the dictionary $\mathcal{D}_N$.



We explored two approaches to threshold the latent features $\mathbf{C}$. In one approach, inspired by \cite{Binary}, we  mapped the features to +1/-1 by incorporating a binarization loss term:
\begin{equation}
\mathcal{L}_{BI} = \|\mathbf{|b|} - \mathbf{1}\|_1
\label{L_BI}
\end{equation}
where, $\mathbf{b} \in \{+1, -1\}^{N}$ and $N$ is the number of bits of the binary code. This module consists of three blocks and two Fully Connected (FC) layers. The first block combines one Conv2D layer with a LeakyRelu followed by  Maxpooling. Then, the last two blocks have the same pattern,  combining Conv2D + LeakyRelu with Avgpooling. The output binary code $\mathbf{b}$ remains the same size as $\mathbf{C}$ but with discrete values.\\

As an alternative approach, we used the Gumbel re-parametrization trick on $\mathbf{C}$, followed by a feature-wise sigmoid function $ \sigma(.)$ to change each element drawn from a Bernouilli distribution, to learn the categorical distribution of $\mathbf{b}$, where $\mathbf{b} \in \{0,1\}^N$. That is, we define:
$\mathbf{g}(\mathbf{C}) \thicksim \text{Bern}(\sigma(\mathbf{\textbf{Gumbel}(|\mathbf{C}|}; \theta))$
where we use absolute value to take care of both positive and negative values, and $\theta$ are the Gumbel parameters. Then, the binarization is done by setting $\mathbf{b}(i) = 1$ if $\mathbf{g}(i) > \alpha$ and $\mathbf{b}(i) = 0$, otherwise,  
where $\alpha$ is the Gumbel threshold. Finally, we used the training loss function: 
\begin{equation}
\mathcal{L}_{Gumbel} = \|\mathbf{b}\|_1
\label{L_gumbel}
\end{equation}

\noindent$\bullet$ {\bf \underline {Classification Head}:} It takes the binary invariant features from the binarization module and outputs the features for the action classifier. It consists of three 1D-Conv blocks(Conv1D+BN+LeakyRelu), two 2D-Conv blocks(Conv2D+BN+LeakyRelu) and one MLP block. The first three 1D Conv-blocks  capture the global and local features of the given input features, while the following two 2D-Conv blocks take the concatenation of global/local features. The MLP block outputs the final action class predictions. This module uses cross entropy to compute the classification loss $\mathcal{L}_{class}$ for action recognition with $c$ classes:
\begin{equation}
\mathcal{L}_{class} = -\sum_{i=1}^{c}t_{i}\log(p_{i}^{rhs})
\label{L_cl}
\end{equation}
where, $t_{i}$ is the true label and $p_{i}$ is the probability of the $i^{th}$ class.
More details of this module can be found in the supplemental material.

\noindent  {\bf \underline{Training Loss}:} The DIR branch is trained with a combination of the modules losses: 
$
\mathcal{L}_{DIR} = \lambda_{1}  \mathcal{L}_{class}+ \lambda_{2}  \mathcal{L}_{B} + \lambda_{3}  \mathcal{L}_{D}$,
 where the binarization loss $\mathcal{L}_B$ is either (\ref{L_BI}) or (\ref{L_gumbel}), depending on which binarization module is used.

\subsection{Enforcing Dependencies between Trajectories}
\begin{figure}[t]
\centering
\includegraphics[height= 1.41in, width=1\linewidth]{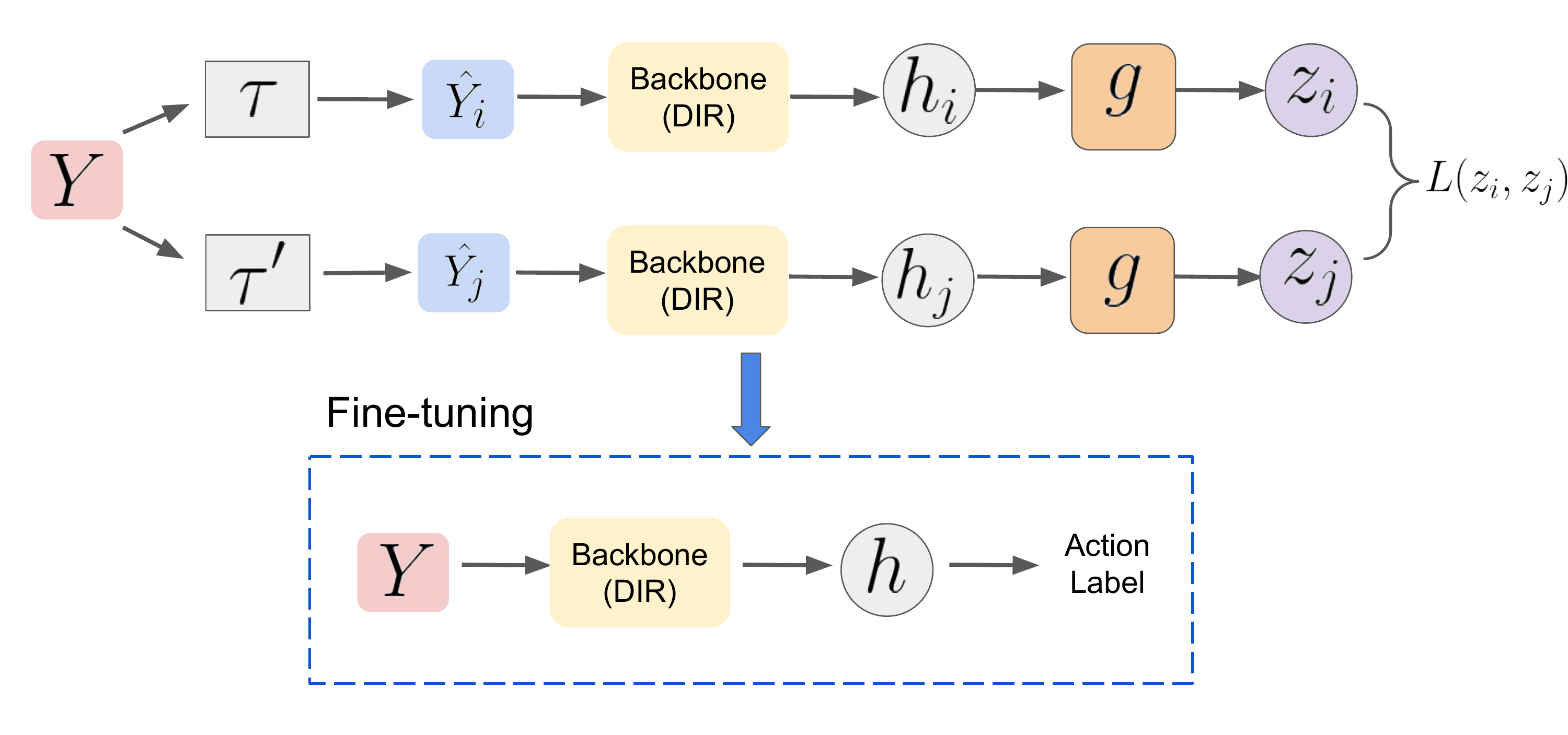}
\caption{\textbf{Training DIR stream with Contrastive Loss.} Assuming '$Y$' represents input a 2D/3D skeleton, $\tau$ and $\tau'$ are two different random affine transformations to augment the data with positve pairs. '$g$' is a projection head  to learn feature projections from the DIR representations $h_{i}$ and $h_{j}$. The contrastive  loss \cite{chen2020simple} is used to maximize the agreement between the projected features $z_{i}$ and $z_{j}$.}
\vspace{-0.55cm}
\label{fig:contrastive}
\end{figure}


A shortcoming of the DIR branch as described above is that equation (\ref{eq:C}) decouples  the  coordinates ($x$, $y$, and $z$) of each joint  and ignores physical couplings between pairs of joints (i.e. shoulder and elbow are connected),  potentially using significantly more poles than strictly needed. To address this issue, we propose  two improvements. Firstly, we propose to  use a contrastive learning strategy similar to \cite{chen2020simple} as illustrated in Fig.~\ref{fig:contrastive}, to encourage the trajectories of the coordinates of each joint to share poles. Here, the positive augmented examples are obtained by applying random affine transformations to the input skeletons before passing them through the DIR branch and a projection head $g(.)$. After training is completed, we throw away the projection head and fine tune the DIR branch. Secondly, to encourage the network to learn that the motion of the joints is constrained by the limbs connecting them, we augment the input data to also include the trajectories of the coordinates of the middle point of each of the limbs. The effectiveness of these approaches is evaluated in our  ablation experiments.

\section{Context Information Representation} \label{sec:CIR}

While the skeletons provide critical view-invariant motion information,  RGB data can also provide useful scene context. Thus, we incorporate an I3D \cite{2017_carreira} based RGB branch to capture a context information representation (CIR)  from  the RGB frames when they are available.  A diagram showing the components of this stream is provided at the bottom of Fig.~\ref{fig:architecture details} and the details are described next. 

{ We modified the original I3D architecture by using a two-branch design to better solve our problem. ROIs are cropped around each actor, resized to $3\times 224\times 224$ and then fed into an `I3D head' to get local features for each actor while another `I3D head' takes raw images to capture global features through  time, simultaneously. Both I3D heads use the same layers as \cite{2017_carreira}: a `conv3-1a-7x7' layer followed by a 3D max pooling layer. The local and global features provided by the heads are concatenated channel-wise to enrich the information from RGB images. Then, they are fed into the `I3D Blocks', consisting of the I3D layers from `conv3-1a' until `mixed-4c'. }Finally, the layer `Temporal pooling', which is a block combining two Conv3D layers, pools features along the temporal domain.

{The loss function used by CIR to  learn the probabilities for each action class  is defined as: }
$\mathcal{L}_{CIR} = -\sum_{i=1}^{c}t_{i}\log(p_{i}^{rgb})$. {The combined probability is defined as $p_{i}^{'} = \mathcal{F}(\beta_{1} f_{i}^{rgb} + \beta_{2} f_{i}^{rhs})$\footnote{In our experiments, we set $\beta_{1}:\beta_{2} = 1:1$}, where $\mathcal{F}$ is the last fusion layer combining features $f_{i}^{rgb}$ and $f_{i}^{rhs}$. Then, the 
 overall loss function using the DIR and CIR streams is given by: }
\begin{equation}
\mathcal{L}_{2-stream} = -\lambda_{1}\sum_{i=1}^{c}t_{i}\log(p_{i}^{'}) + \lambda_{2}  \mathcal{L}_{B} + \lambda_{3}  \mathcal{L}_{D}
\label{L_2stream}
\end{equation}

\begin{table*}[t]
\vspace{-.5cm}
\centering
\caption{\textbf{Ablation study on different architecture configurations and sampling strategies.} Input data is \textbf{only RGB video}. `Baseline' uses a vanilla DYAN encoder \cite{2018_liu_dyan} without binarization,  `BI'  and `Gumbel' indicate the type  Binarization Module, and `*I3D' stands for our modified version from the original paper\cite{2017_carreira}. }
\label{abl:architecture}
\centering
\begin{adjustbox}{width=\textwidth}
    \begin{tabular}{|ccccccccccccc|}
    \hline
\multicolumn{13}{|c|}{Ablation Study: N-UCLA Cross View}                                                                                                                                       \\ \hline
\multicolumn{1}{|l|}{}             & \multicolumn{8}{c|}{DIR Stream}                                                                                                                                                                                                                                                       & \multicolumn{2}{c|}{CIR Stream}                                  & \multicolumn{2}{c|}{DIR + CIR Streams}                              \\ \hline
\multicolumn{1}{|c|}{Architecture} & \multicolumn{1}{c|}{Baseline}    & \multicolumn{1}{c|}{RHS}       & \multicolumn{1}{c|}{RHS+BI}    & \multicolumn{1}{c|}{RHS+Gumbel}  & \multicolumn{1}{c|}{Baseline}    & \multicolumn{1}{c|}{RHS}       & \multicolumn{1}{c|}{RHS+BI}    & \multicolumn{1}{c|}{RHS+Gumbel}  & \multicolumn{1}{c|}{*I3D}   & \multicolumn{1}{c|}{*I3D}          & \multicolumn{1}{c|}{RHS+BI+*I3D}     & RHS+Gumbel+*I3D \\ \hline
\multicolumn{1}{|c|}{Sampling}     & \multicolumn{1}{c|}{Single}      & \multicolumn{1}{c|}{Single}      & \multicolumn{1}{c|}{Single}      & \multicolumn{1}{c|}{Single}        & \multicolumn{1}{c|}{Multiple}    & \multicolumn{1}{c|}{Multiple}    & \multicolumn{1}{c|}{Multiple}    & \multicolumn{1}{c|}{Mutiple}       & \multicolumn{1}{c|}{Single} & \multicolumn{1}{c|}{Multiple}      & \multicolumn{1}{c|}{Multiple}          & \multicolumn{1}{c|}{Multiple}           \\ \hline
\multicolumn{1}{|c|}{Accuracy(\%)} & \multicolumn{1}{c|}{86.0}        & \multicolumn{1}{c|}{86.7}        & \multicolumn{1}{c|}{89.0}        & \multicolumn{1}{c|}{{90.2}} & \multicolumn{1}{c|}{87.1}        & \multicolumn{1}{c|}{87.5}        & \multicolumn{1}{c|}{90.1}        & \multicolumn{1}{c|}{\textbf{\textcolor{red}{92.9}}} & \multicolumn{1}{c|}{87.5}   & \multicolumn{1}{c|}{\textbf{\textcolor{red}{91.2}}} & \multicolumn{1}{c|}{94.4}              & \textbf{\textcolor{red}{95.7}}     \\ \hline
            \hline
        \multicolumn{13}{|c|}{Ablation Study: NTU-RGB+D 60 Cross View}                                                                           \\ \hline
\multicolumn{1}{|l|}{}             & \multicolumn{8}{c|}{DIR Stream}                                                                                                                                                                                                                                                       & \multicolumn{2}{c|}{CIR Stream}                                  & \multicolumn{2}{c|}{DIR + CIR Streams}                              \\ \hline
\multicolumn{1}{|c|}{Architecture} & \multicolumn{1}{c|}{Baseline}    & \multicolumn{1}{c|}{RHS}       & \multicolumn{1}{c|}{RHS+BI}    & \multicolumn{1}{c|}{RHS+Gumbel}  & \multicolumn{1}{c|}{Baseline}    & \multicolumn{1}{c|}{RHS}       & \multicolumn{1}{c|}{RHS+BI}    & \multicolumn{1}{c|}{RHS+Gumbel}  & \multicolumn{1}{c|}{*I3D}   & \multicolumn{1}{c|}{*I3D}          & \multicolumn{1}{c|}{RHS+BI+*I3D}     & RHS+Gumbel+*I3D \\ \hline
\multicolumn{1}{|c|}{Sampling}     & \multicolumn{1}{c|}{Single}      & \multicolumn{1}{c|}{Single}      & \multicolumn{1}{c|}{Single}      & \multicolumn{1}{c|}{Single}        & \multicolumn{1}{c|}{Multiple}    & \multicolumn{1}{c|}{Multiple}    & \multicolumn{1}{c|}{Multiple}    & \multicolumn{1}{c|}{Mutiple}       & \multicolumn{1}{c|}{Single} & \multicolumn{1}{c|}{Multiple}      & \multicolumn{1}{c|}{Multiple}          & Multiple          \\ \hline
    \multicolumn{1}{|c|}{Accuracy(\%)} & \multicolumn{1}{c|}{83.3}        & \multicolumn{1}{c|}{84.8}        & \multicolumn{1}{c|}{87.6}        & \multicolumn{1}{c|}{{89.5}} & \multicolumn{1}{c|}{84.1}        & \multicolumn{1}{c|}{85.8}        & \multicolumn{1}{c|}{90.0}        & \multicolumn{1}{c|}{\textbf{\textcolor{red}{91.3}}} & \multicolumn{1}{c|}{84.7}   & \multicolumn{1}{c|}{\textbf{\textcolor{red}{90.2}}} & \multicolumn{1}{c|}{93.1}              & \textbf{\textcolor{red}{95.0} }    \\ \hline
    \end{tabular}
\end{adjustbox}

\end{table*}

\begin{table}[]
\centering
\caption{\textbf{Ablation study on  training strategies.} This experiment evaluates the effect of different learning strategies. `DIR' and `DIR+CIR' stand for when there are no additional learning strategies. `CL' stands for contrastive learning. The input is RGB.}
\label{abl:contrastive}
\begin{adjustbox}{width=\linewidth}
\begin{tabular}{|cccccc|}
\hline
\multicolumn{6}{|c|}{Impact of Learning Strategies}                                                                                                                                                                                                       \\ \hline
\multicolumn{2}{|c|}{}                                                                                                                                           & \multicolumn{2}{c|}{N-UCLA}                           & \multicolumn{2}{c|}{NTU-60}      \\ \cline{3-6} 
\multicolumn{2}{|c|}{}                                                                                                                                           & \multicolumn{1}{c|}{CV}   & \multicolumn{1}{c|}{CS}   & \multicolumn{1}{c|}{CV}   & CS   \\ \hline

\multicolumn{1}{|c|}{\multirow{3}{*}{Baseline\cite{2018_liu_dyan}}}     & \multicolumn{1}{c|}{Baseline}                                                                                 & \multicolumn{1}{c|}{87.1} & \multicolumn{1}{c|}{85.4} & \multicolumn{1}{c|}{84.1} & 83.3 \\ \cline{2-6} 
\multicolumn{1}{|c|}{}                                & \multicolumn{1}{c|}{Pre-training}                                                                    & \multicolumn{1}{c|}{89.2} & \multicolumn{1}{c|}{87.9} & \multicolumn{1}{c|}{86.5} & 84.7 \\ \cline{2-6} 
\multicolumn{1}{|c|}{}                                & \multicolumn{1}{c|}{\begin{tabular}[c]{@{}c@{}}Pre-training+CL\\ 
\end{tabular}} & \multicolumn{1}{c|}{90.4} & \multicolumn{1}{c|}{88.9} & \multicolumn{1}{c|}{87.3} & 86.0 \\ \hline \hline

\multicolumn{1}{|c|}{\multirow{3}{*}{DIR Stream}}     & \multicolumn{1}{c|}{DIR}                                                                                 & \multicolumn{1}{c|}{92.9} & \multicolumn{1}{c|}{91.5} & \multicolumn{1}{c|}{91.3} & 90.0 \\ \cline{2-6} 
\multicolumn{1}{|c|}{}                                & \multicolumn{1}{c|}{Pre-trainingRHS}                                                                    & \multicolumn{1}{c|}{93.3} & \multicolumn{1}{c|}{93.1} & \multicolumn{1}{c|}{93.0} & 90.9 \\ \cline{2-6} 
\multicolumn{1}{|c|}{}                                & \multicolumn{1}{c|}{\begin{tabular}[c]{@{}c@{}}Pre-training RHS+CL\\
\end{tabular}} & \multicolumn{1}{c|}{96.6} & \multicolumn{1}{c|}{94.5} & \multicolumn{1}{c|}{97.3} & 93.1 \\ \hline \hline
\multicolumn{1}{|c|}{\multirow{3}{*}{DIR+CIR Stream}} & \multicolumn{1}{c|}{DIR+CIR}                                                                             & \multicolumn{1}{c|}{95.7} & \multicolumn{1}{c|}{92.3} & \multicolumn{1}{c|}{95.0} & 92.5 \\ \cline{2-6} 
\multicolumn{1}{|c|}{}                                & \multicolumn{1}{c|}{Pre-training RHS}                                                                    & \multicolumn{1}{c|}{97.4} & \multicolumn{1}{c|}{94.6} & \multicolumn{1}{c|}{97.2} & 94.9 \\ \cline{2-6} 
\multicolumn{1}{|c|}{}                                & \multicolumn{1}{c|}{\begin{tabular}[c]{@{}c@{}}Pre-training RHS+CL\\
\end{tabular}} & \multicolumn{1}{c|}{\textbf{\textcolor{red}{98.6}}} & \multicolumn{1}{c|}{\textbf{\textcolor{red}{96.0}}} & \multicolumn{1}{c|}{\textbf{\textcolor{red}{99.1}}} & \textbf{\textcolor{red}{97.2}} \\ \hline
\end{tabular}
\end{adjustbox}
\end{table}

\begin{table}[]
\centering
\caption{\textbf{Ablation on DIR Input Sources.} `J',  `J*', and `J$_{gt}$' indicate the  source of the 2D joints from RGB data: \cite{2017_cao_realtime}, \cite{2022_duan_revisiting}, and ground truth, respectively.  `J+L' stands for joint and limb data. Followed with \cite{2022_duan_revisiting}, there are eight limb keypoints. }
\label{diff.input}
\begin{adjustbox}{width=\linewidth}
    \begin{tabular}{|ccccccc|}
\hline
\multicolumn{7}{|c|}{Input Variations on NTU-60}                                                                                                                                                                                                                             \\ \hline
\multicolumn{2}{|c|}{}                                                                                                              & \multicolumn{1}{c|}{\# of joints+ limbs} & \multicolumn{1}{c|}{CV}   & \multicolumn{1}{c|}{CS}   & \multicolumn{1}{c|}{FLOPS(G)} & \#Params(M) \\ \hline
\multicolumn{1}{|c|}{\multirow{2}{*}{PoseConv3D \cite{2022_duan_revisiting}}}                                              & \multicolumn{1}{c|}{J*}            & \multicolumn{1}{c|}{17}          & \multicolumn{1}{c|}{96.6} & \multicolumn{1}{c|}{93.7} & \multicolumn{1}{c|}{15.90}    & 2.00        \\ \cline{2-7} 
\multicolumn{1}{|c|}{}                                                                         & \multicolumn{1}{c|}{J*+L}          & \multicolumn{1}{c|}{17+8}        & \multicolumn{1}{c|}{97.1} & \multicolumn{1}{c|}{94.1} & \multicolumn{1}{c|}{-}        & -           \\ \hline \hline
\multicolumn{1}{|c|}{\multirow{4}{*}{\begin{tabular}[c]{@{}c@{}}Ours\\ (CL-DIR)\end{tabular}}} & \multicolumn{1}{c|}{J*}            & \multicolumn{1}{c|}{17}          & \multicolumn{1}{c|}{96.8} & \multicolumn{1}{c|}{92.9} & \multicolumn{1}{c|}{\bf 9.80}     & {\bf 1.16}        \\ \cline{2-7} 
\multicolumn{1}{|c|}{}                                                                         & \multicolumn{1}{c|}{J}             & \multicolumn{1}{c|}{25}          & \multicolumn{1}{c|}{97.3} & \multicolumn{1}{c|}{93.1} & \multicolumn{1}{c|}{9.84}     & 1.19        \\ \cline{2-7} 
\multicolumn{1}{|c|}{}                                                                         & \multicolumn{1}{c|}{J*+L}          & \multicolumn{1}{c|}{17+8}        & \multicolumn{1}{c|}{98.3} & \multicolumn{1}{c|}{94.5} & \multicolumn{1}{c|}{10.51}    & 1.21        \\ \cline{2-7} 
\multicolumn{1}{|c|}{}                                                                         & \multicolumn{1}{c|}{J+L}           & \multicolumn{1}{c|}{25+8}        & \multicolumn{1}{c|}{\textcolor{red}{\bf 98.4}} & \multicolumn{1}{c|}{\textcolor{red}{\textbf{94.5}}} & \multicolumn{1}{c|}{11.00}    & 1.38        \\ \hline \hline
\multicolumn{1}{|c|}{\multirow{2}{*}{\begin{tabular}[c]{@{}c@{}}Ours\\ (CL-DIR)\end{tabular}}} & \multicolumn{1}{l|}{J$_{gt}$}     & \multicolumn{1}{c|}{20}          & \multicolumn{1}{l|}{98.1} & \multicolumn{1}{l|}{93.7} & \multicolumn{1}{c|}{9.90}      & 1.18        \\ \cline{2-7} 
\multicolumn{1}{|c|}{}                                                                         & \multicolumn{1}{l|}{J$_{gt}$ + L} & \multicolumn{1}{c|}{20+8}        & \multicolumn{1}{l|}{\textcolor{red}{\textbf{99.0}}} & \multicolumn{1}{l|}{\textcolor{red}{\textbf{95.2}}} & \multicolumn{1}{c|}{10.2}     & 1.24        \\ \hline
\end{tabular}

\end{adjustbox}
\end{table}

\section{Sampling Strategies}
The backbone of the network uses a Sampling Clip module to  process shorter sequences.  We explored two possible sampling strategies, which are  described next.

\noindent\textbf{Multi-clips. }Consider the input image sequence $\mathcal{I}_{1:L}$ and its skeleton sequence $\mathcal{X}_{1:L}$, where $L$ is the total length of the input. We first uniformly sample $n$ anchor frames from the sequence and extract $t$ frames centered at each of these anchor frames.  For instance, if the first anchor is the $j^{th}$ input frame, the first image clip $\mathcal{I}_{t,1}$ is made of frames $\mathcal{I}_{j-\frac{t}{2}:j+\frac{t}{2}}$.
Therefore, $\mathcal{I}_{1:L}$ is sampled to $ \{I_{t,1}, I_{t,2},..., I_{t,n}\}$ and the corresponding skeleton sequences $\mathcal{X}_{1:L}$ are sampled to $\{X_{t,1}, X_{t,2},..., X_{t,n}\}$. Note that these clips may or may not overlap.  The network learns the  representation from each clip and outputs the final decision by combining all clips together: \textit{Action Label}  $= \mathbf{{\arg\max}}(\frac{1}{n}\sum_{i=1}^{n} P_{i})$
where, $P_{i}$ is the combined probability for the $i^{th}$ clip.

\noindent\textbf{Single-clip.} 
Alternatively,  we tested sub-sampling the sequence into a single clip. Here, the sampled clip consists of only the uniformly sampled anchor frames. In this case, the action label is given by \textit{Action Label} $= {\arg\max}(P)$
where $P$ is the final probability from the network.

\section{Reproducibility and Implementation Details}\label{sec:implementation}

 A Pytorch implementation of our approach  will be made available. Pseudo code  is also provided in the supplemental material. 
 The input skeletons  were  normalized  by the mean and variance, which were computed over the entire training sets. We also resized the input images to 3x224x224 and normalized them using the mean (0.485,0.456,0.406) and the standard deviation (0.229,0.224,0.225). We use SGD optimizer and  set the learning rate to  1e-4 for the RHS module and to 1e-3 for the rest of the modules (e.g classifier).

The hyper-parameter $\lambda$ in (\ref{Ldyan}) was chosen by  using a greedy search between 0.1 and 100 and balancing reconstruction error versus sparsity. In the end,  if training only the DIR branch,  we set $\lambda=0.2$ in (\ref{Ldyan}), $\lambda_1:\lambda_2:\lambda_3=$2:1:0.1 in the loss ${\cal L}_{DIR}$, and the Gumbel threshold to 0.51; if training DIR and CIR, we set $\lambda=0.1$, $\lambda_1:\lambda_2:\lambda_3=$1:1:0.1 in (\ref{L_2stream}), and the Gumbel threshold to 0.505. The Gumbel threshold was determined by drawing the distribution of dynamic representations across the entire training set.  During inference, the Gumbel threshold was kept the same. Furthermore, since the binarization loss term is unsupervised in the sense that its  ground truth is unknown, we found beneficial to pre-train a standalone binarization module using synthetic data and fine tune the pre-trained during the end-to-end training.
 

\begin{table*}[h]
\centering
\caption{\textbf{Comparison of all setups on UWA3DII dataset}. RGB input modality.}

\label{tab:UWA3D}
\begin{adjustbox}{width=1.0\textwidth}
\begin{tabular}{|lllllllllllllc|}
\hline
\multicolumn{14}{|c|}{Accuracy(\%) on the UWA3D dataset}                                                                                                                                                                    \\ \hline
\multicolumn{1}{|l|}{Training views}                & \multicolumn{2}{l|}{V1\&V2}                                             & \multicolumn{2}{l|}{V1\&V3}                                             & \multicolumn{2}{l|}{V1\&V4}                                             & \multicolumn{2}{l|}{V2\&V3}                                             & \multicolumn{2}{l|}{V2\&V4}                                             & \multicolumn{2}{l|}{V3\&V4}                                             &  Average \\ \hline
\multicolumn{1}{|l|}{Testing views}                 & \multicolumn{1}{l|}{V3}            & \multicolumn{1}{l|}{V4}            & \multicolumn{1}{l|}{V2}            & \multicolumn{1}{l|}{V4}            & \multicolumn{1}{l|}{V2}            & \multicolumn{1}{l|}{V3}            & \multicolumn{1}{l|}{V1}            & \multicolumn{1}{l|}{V4}            & \multicolumn{1}{l|}{V1}            & \multicolumn{1}{l|}{V3}            & \multicolumn{1}{l|}{V1}            & \multicolumn{1}{l|}{V2}            &                          \\ \hline
\multicolumn{1}{|l|}{VA-fusion\cite{VA-fusion}}              & \multicolumn{1}{l|}{80.9}          & \multicolumn{1}{l|}{84.3}          & \multicolumn{1}{l|}{78.7}          & \multicolumn{1}{l|}{86.2} & \multicolumn{1}{l|}{75.2}          & \multicolumn{1}{l|}{73.3}          & \multicolumn{1}{l|}{87.6} & \multicolumn{1}{l|}{84.3}          & \multicolumn{1}{l|}{86.0}          & \multicolumn{1}{l|}{74.9}          & \multicolumn{1}{l|}{86.4}          & \multicolumn{1}{l|}{79.5}          & 81.4                     \\ \hline
\multicolumn{1}{|l|}{VT+GARN\cite{2021_huang_view}}                & \multicolumn{1}{l|}{79.5}          & \multicolumn{1}{l|}{83.4}          & \multicolumn{1}{l|}{75.3}          & \multicolumn{1}{l|}{85.2}          & \multicolumn{1}{l|}{74.3}          & \multicolumn{1}{l|}{\bf{84.7}} & \multicolumn{1}{l|}{86.3}          & \multicolumn{1}{l|}{84.8}          & \multicolumn{1}{l|}{86.1}          & \multicolumn{1}{l|}{75.5}          & \multicolumn{1}{l|}{86.4}          & \multicolumn{1}{l|}{74.1}          & 81.3                     \\ \hline \hline
\multicolumn{1}{|l|}{\textbf{Ours} (CL-DIR+CIR)} & \multicolumn{1}{l|}{\textbf{84.2}} & \multicolumn{1}{l|}{\textbf{86.9}} & \multicolumn{1}{l|}{\textbf{80.8}} & \multicolumn{1}{l|}{\textbf{87.1}}          & \multicolumn{1}{l|}{\textbf{77.7}} & \multicolumn{1}{l|}{80.2}          & \multicolumn{1}{l|}{\textbf{88.3}}          & \multicolumn{1}{l|}{\textbf{87.9}} & \multicolumn{1}{l|}{\textbf{88.5}} & \multicolumn{1}{l|}{\textbf{80.1}} & \multicolumn{1}{l|}{\textbf{88.9}} & \multicolumn{1}{l|}{\textbf{82.7}} & \textcolor{red}{\textbf{84.4}}            \\ \hline
\end{tabular}
 \end{adjustbox}

\end{table*}

\vspace{-0.2cm}
\section{Experiments}
\vspace{-0.1cm}
We performed experiments using four benchmark datasets for CVAR (N-UCLA, NTU-RGB+D60, NTU-RGB+D 120, and UWA3D Multiview II) and one dataset for single view action detection (sub-JHMDB). These datasets are described in detail in  the supplemental material.

\subsection{Ablation Studies}
We conducted ablation studies on the N-UCLA and NTU 60 datasets, Cross-view(CV) setup, to  evaluate the effectiveness of each component of our approach. Comparisons are made against a baseline  vanilla DYAN \cite{2018_liu_dyan} encoder.


\noindent {\bf Architecture Variations and Sampling Strategies.} Table~\ref{abl:architecture} shows that  each of the proposed modules, {RHS},  binarization, and sampling increases performance. The largest improvements are observed when adding binarization and multiple sampling. We believe that the contribution of the binarization modules is to correctly identify the invariant features, and that using multiple clipping ensures that each clip captures well these invariants.  The experiments also show that using the  DIR stream alone has better performance than using the CIR stream alone, highlighting the benefits of using invariance. However, the two streams bring complementary features since using them together improves the overall performance.

\noindent {\bf Training Strategies.} We evaluated the effect of  pre-training the  RHS module and of using contrastive learning to train  the DIR branch. Here, pre-training means that the RHS dictionary is pre-trained on the input reconstruction loss. Table~\ref{abl:contrastive} shows that both strategies are beneficial, with contrastive learning providing the largest boost.

\noindent{\bf DIR Input Data.} We evaluated the impact of using different skeleton input sources as well as the number of input sequences used on classification performance and computational costs. For input sources, we considered 2D skeletons from RGB provided by \cite{2022_duan_revisiting}, computed with Openpose \cite{2017_cao_realtime}  and ground truth. Each of these sources provide a different number of joints. In addition, we evaluated the effect of adding sequences for the mid point of the limbs. A summary of these experiments is given in Table~\ref{diff.input}. The performances using either of the pose detectors are very similar, marginally better when using Openpose. Using ground truth skeletons also provides a bit of improvement. In all cases, adding limb data boosts performance. Finally, the average FLOPS and number of parameters  are 10G and 1.23M, respectively. In comparison, the previous SOTA uses 15.9G FLOPS and 2M parameters.


\noindent {\bf Additional ablation studies.}  A summary of these experiments are included in the supplemental material: (1) We evaluated the benefits  of using a re-weighted heuristic in conjunction with FISTA in the DIR stream,  pre-training the binarization modules, and
 fusing the DIR and CIR streams.  
(2) The common protocol for cross-view on the N-UCLA dataset, calls for training on views 1 and 2 and testing on 3. We tested the performance of the proposed approach using all possible combinations training with two views and testing with the remaining one.  This experiment showed that  view 1 is the most challenging set up.  We hypothesize that this is because view 1 has significant perspective distortion 
and our approach assumes affine invariance.

\subsection{Comparisons against the SOTA}

We compared the performance of our  architecture using multiple clipping, RHS, Gumbel binarization and the CIR stream (if using RGB data), against SOTA using different input modalities: RGB alone, 3D skeletons alone, and RGB together with 3D skeletons.
For fair comparison, when comparing against RGB approaches, the DIR stream does not use the available skeleton ground truth information. Instead, it uses as input 2D skeletons detected using Openpose\cite{2017_cao_realtime} on the given videos.  When comparing against 3D approaches, the input to the DIR stream is the same as used by other approaches, i.e.  the skeletons provided in the datasets. For 3D approaches, we reported performance with and without using the CIR stream. We also tested using 3D skeletons estimated from RGB videos  \cite{videoPose3D}. However, (see Table~\ref{tab:NUCLA}), the skeletons are not accurate and performance suffered.   As is traditional in the literature, in addition to the Cross-view (CV) setup, we also evaluated our approach by following the Cross-subject (CS) protocol for all datasets. 


{The results of these experiments are reported in Tables \ref{tab:UWA3D}, \ref{tab:NUCLA},  \ref{tab:ntu60}, and \ref{tab:NTU120}. Our approach consistently improves the CVAR SOTA on all four datasets,  regardless of the input modality used  (RGB alone, 3D skeletons alone, and RGB and 3D skeletons together). The largest improvements are observed when restricting the input data to RGB videos, with performance achieving comparable levels to the performance  using 3D data. Indeed, our approach reduced the 2D-3D performance gap to 0.5\%, 0.3\% and 1.9\%  in the N-UCLA, NTU 60, and NTU 120 datasets. These experiments also show the flexibility of our architecture, since it can be used with different types of input modalities with minimal changes. Even though the proposed architecture was not designed for the cross-subject task, our experiments show that the proposed architecture outperforms the SOTA in this task for the N-UCLA, NTU-60, and NTU-120 using all input modalities. }

 Finally, we also tested  our approach on single view action recognition with the sub-JHMDB dataset.  The results of this experiment are summarized in the supplemental material. Our approach achieved 92.5\% accuracy using the DIR and CIR streams, outperforming the current SOTA.

\begin{table}[h]
\caption{\textbf{Comparison against SOTA Cross-Subject (CS) and Cross-View (CV) on N-UCLA.}  A \textbf{$\dagger$} before a method indicates that the performance  is from \cite{2020_vyas_multiview}.  `3D Skeleton*' indicates that the 3D skeletons were obtained from RGB videos using \cite{videoPose3D}, while `3D Skeleton' is from ground truth data.}\label{tab:NUCLA}
\vspace{-1em}
\begin{center}
\begin{adjustbox}{width=\linewidth}
\begin{tabular}{ |l|c|c|c|  }
\hline\multicolumn{4}{|c|}{Accuracy(\%) on N-UCLA}\\
 \hline
Method & Modality & CS & CV\\
  \hline
 $\dagger$Hanklets\cite{2012_li_cross}   & RGB & 54.2 & 45.2 \\
 $\dagger$DV-Views\cite{2012_li_discriminative}  & RGB  & 50.7 & 58.5 \\
 $\dagger$LRCN\cite{2015_donahue_long}       & RGB    & - & 64.7\\
 $\dagger$nCTE\cite{2014_gupta_3d}       & RGB    & -  & 68.6\\
 UMVRL\cite{2020_vyas_multiview} & RGB  & 87.5 & 83.1 \\ \hline
 \textbf{Ours}(CIR)             & RGB  & \textcolor{red}{\textbf{90.9}} & \textcolor{red}{\textbf{91.2} }\\
 \hline \hline
  VPN++\cite{2021_das_vpn++}      & RGB+(J) & - & 91.9 \\ \hline
 \textbf{Ours} (CL-DIR+CIR)      & RGB+(J) & {96.0} & {98.6} \\
 \textbf{Ours}(CL-DIR+CIR)     & RGB+(J+L) &  \textcolor{red}{\textbf{97.5} } & \textcolor{red}{\textbf{99.4}}\\
 \hline\hline
 ESV\cite{2017_liu_enhanced}  & 3D Skeleton & -  & 92.6\\
 `TS+SS'C\cite{2021_yang_skeleton}  & 3D Skeleton & -  & 94.0\\
 VA-fusion\cite{2019_zhang_view} & 3D Skeleton & - & 95.3\\
 CTR-GCN\cite{2021_chen_channel}  & 3D Skeleton & -  & 96.5\\
\hline
\textbf{Ours}(CL-DIR) & 3D Skeleton*  &     90.0                   & 91.4\\
\textbf{Ours} (CL-DIR)  & 3D Skeleton & 96.2 & 98.5 \\
\textbf{Ours} (CL-DIR)  & 3D Skeleton+L & \textcolor{red}{\textbf{97.3}} & \textcolor{red}{\textbf{98.9}} \\
\hline \hline
$\dagger$MST-AOG\cite{2014_wang_cross} & RGB+3D Skeleton & 81.6 & 73.3 \\
 $\dagger$NKTM\cite{2015_rahmani_learning} & RGB+3D Skeleton & - & 75.8 \\
  VPN++\cite{2021_das_vpn++} & RGB+3D Skeleton & - & 93.5 \\ \hline
\textbf{Ours} (CL-DIR+CIR) & RGB+3D Skeleton* & 92.6 & 93.5 \\
\textbf{Ours} (CL-DIR+CIR) & RGB+3D Skeleton & {{97.9}} & {{99.8}} \\
\textbf{Ours} (CL-DIR+CIR) & RGB+(3D Skeleton+L) &\textcolor{red}{\textbf{98.7}} & \textcolor{red}{\textbf{99.9}} \\
\hline
\end{tabular}
\end{adjustbox}
\end{center}
\end{table}

\begin{table}
\caption{\textbf{Comparison against SOTA Cross-Subject (CS) and Cross-View (CV) on  NTU-60.}  Note that for \cite{2020_davoodikakhki_hierarchical}, 2D skeletons is projected from ground truth 3D skeletons.}
\label{tab:ntu60}
\centering
\begin{adjustbox}{width=\linewidth}
\begin{tabular}{cc}
\begin{tabular}{ |l||c|c|c|  }
\hline\multicolumn{4}{|c|}{Accuracy(\%) on NTU-60}\\
 \hline
Method & Modality & CS & CV\\
 \hline
 CNN-LSTM\cite{2017_luo_unsupervised}  & RGB & 56.0 & -    \\
 DA-NET\cite{2018_wang_dividing}    & RGB & -    & 75.3 \\
 Att-LSTM\cite{2018_zhang_adding}  & RGB & 63.3 & 70.6 \\
 CNN-BiLSTM\cite{2018_li} & RGB & 55.5 & 49.3 \\
 UMVRL\cite{2020_vyas_multiview}     & RGB & 82.3 & 86.3 \\
 \hline
 \textbf{Ours}(CIR)          &RGB & \textcolor{red}{\bf 89.7} & \textcolor{red}{\bf 90.2} \\
 \hline \hline
  HNCNP\cite{2020_davoodikakhki_hierarchical}    & RGB+J$_{gt'}$ & 95.7 & 98.8 \\
 PoseConv3D\cite{2022_duan_revisiting}  & RGB+(J*+L) & 97.0 & \textcolor{red}{\bf 99.6} \\
 \hline
 \textbf{Ours}(CL-DIR+CIR) & RGB+(J*) &  97.2  & 99.0 \\
 \textbf{Ours}(CL-DIR+CIR) & RGB+(J*+L) &  {97.5}    & {\bf 99.4} \\
 \textbf{Ours} (CL-DIR+CIR) & RGB+(J) & 97.2 & 99.1 \\
 \textbf{Ours}(CL-DIR+CIR) &RGB+(J+L) &    \textcolor{red}{\textbf{97.6}}         & {\bf 99.4} \\
 
 \hline
 \hline
 GeomNet\cite{2021_nguyen_geomnet}         & 3D Skeleton & 93.6 & 96.3 \\
 Else-Net\cite{2021_li_else}         & 3D Skeleton& 91.6 & 96.4 \\
 CTR-GCN\cite{2021_chen_channel}         & 3D Skeleton & 92.4 & 96.8 \\
ACFL-CTR-GCN\cite{2022_wang_skeleton}     & 3D Skeleton & 92.5 & 97.1 \\
PYSKL\cite{2022_duan_pyskl}    & 3D Skeleton & 92.6 & 97.4 \\
 KShapeNet\cite{2021_friji_geometric}         & 3D Skeleton & {97.0} & 98.5 \\
 \hline
  \textbf{Ours}    (CL-DIR)     & 3D Skeleton & 96.8 & 99.6 \\
  \textbf{Ours}    (CL-DIR)     & 3D Skeleton + L& \textcolor{red}{\bf 97.5} & \textcolor{red}{\bf 99.8} \\
 \hline \hline
 VPN++\cite{2021_das_vpn++}  & RGB+3D Skeleton & 96.6 & 99.1 \\
 \textbf{Ours} (CL-DIR+CIR)        & RGB+3D Skeleton& 97.7 & 99.8 \\
 \textbf{Ours} (CL-DIR+CIR)        & RGB+(3D Skeleton+L)&  \textcolor{red}{\textbf{98.0}}& \textcolor{red}{\textbf{99.9}} \\
 \hline
 \multicolumn{4}{}\\
\end{tabular} 
\end{tabular}
\end{adjustbox}
\end{table}
\vspace{-0.8cm}

\begin{table}[h]
\centering
\caption{\textbf{Comparison against SOTA  Cross-Subject (CS), Cross-Setup(C-setup) on NTU-120.}}
\label{tab:NTU120}
\begin{adjustbox}{width=\linewidth}
\begin{tabular}{|cccc|}
\hline
\multicolumn{4}{|c|}{ Accuracy(\%) on NTU-120}                                                             \\ \hline
\multicolumn{1}{|c|}{Method}              & \multicolumn{1}{c|}{Modality}          & \multicolumn{1}{c|}{CS} & C-setup \\ \hline
\hline
\multicolumn{1}{|c|}{PoseConv3D\cite{2022_duan_revisiting}} & \multicolumn{1}{c|}{RGB + (J+L)} & \multicolumn{1}{c|}{95.3}   &     96.4   \\ 
\hline
\multicolumn{1}{|c|}{\textbf{Ours} (CL-DIR + CIR)} & \multicolumn{1}{c|}{RGB + (J*) } & \multicolumn{1}{c|}{94.2}   &    96.6    \\
\multicolumn{1}{|c|}{\textbf{Ours} (CL-DIR + CIR)} & \multicolumn{1}{c|}{RGB + (J+L) } & \multicolumn{1}{c|}{95.0}   &    96.7    \\
\multicolumn{1}{|c|}{\textbf{Ours} (CL-DIR + CIR)} & \multicolumn{1}{c|}{RGB + (J*+L) } & \multicolumn{1}{c|}{\textcolor{red}{\textbf{95.8}}}   &    \textcolor{red}{ \textbf{97.3}}    \\
\hline
\hline
\multicolumn{1}{|c|}{GeomNet\cite{2021_nguyen_geomnet}}             & \multicolumn{1}{c|}{3D Skeleton}       & \multicolumn{1}{c|}{86.5}   &   87.6      \\ 
\multicolumn{1}{|c|}{CTR-GCN\cite{2021_chen_channel}}             & \multicolumn{1}{c|}{3D Skeleton}       & \multicolumn{1}{c|}{88.9}   &   90.6      \\ 
\multicolumn{1}{|c|}{PYSKL\cite{2022_duan_pyskl}}             & \multicolumn{1}{c|}{3D Skeleton}       & \multicolumn{1}{c|}{88.6}   &   90.8      \\
\multicolumn{1}{|c|}{ACFL-CTR-GCN\cite{2021_dai_learning}}             & \multicolumn{1}{c|}{3D Skeleton}       & \multicolumn{1}{c|}{89.7}   &   90.9      \\
\multicolumn{1}{|c|}{KShapeNet\cite{2021_friji_geometric}}           & \multicolumn{1}{c|}{3D Skeleton}       & \multicolumn{1}{c|}{90.6}   & 86.7        \\ \hline
\multicolumn{1}{|c|}{\textbf{Ours} (CL-DIR)}  & \multicolumn{1}{c|}{3D Skeleton}       & \multicolumn{1}{c|}{{92.7}}   &    {93.5}    \\ 
\multicolumn{1}{|c|}{\textbf{Ours} (CL-DIR)}  & \multicolumn{1}{c|}{3D Skeleton + L}       & \multicolumn{1}{c|}{\textcolor{red}{\textbf{93.6}}}   &    \textcolor{red}{\textbf{95.0}}     \\ 
\hline 
 \hline
\multicolumn{1}{|c|}{VPN++\cite{2021_das_vpn++}} & \multicolumn{1}{c|}{RGB + 3D Skeleton}  & \multicolumn{1}{c|}{90.7}   &     92.5   \\ \hline
\multicolumn{1}{|c|}{\textbf{Ours} (CL-DIR + CIR)} & \multicolumn{1}{c|}{RGB + 3D Skeleton } & \multicolumn{1}{c|}{{96.8} }  &     {98.0}   \\ 
\multicolumn{1}{|c|}{\textbf{Ours} (CL-DIR + CIR)} & \multicolumn{1}{c|}{RGB + (3D Skeleton+L) } & \multicolumn{1}{c|}{\textcolor{red}{\bf 97.7} }  &     \textcolor{red}{\bf 99.2 }   \\ 
\hline

\end{tabular}
\end{adjustbox}
\end{table}

    

\vspace{-0.3cm}
\section{Conclusions}
We introduced a two stream architecture  that learns  dynamics-based invariant features and context features for cross-view  action recognition. The proposed framework is flexible and can be used with different types of input modalities: RGB, 3D Skeletons, or both. Our extensive ablation studies show that both streams contribute to boosting the performance. Comparisons of the proposed approach against the current state of the art methods, using four widely used benchmark datasets, show that our approach outperforms the state of the art in all input modalities and has closed significantly the existing performance gap between RGB and 3D skeleton based approaches.
We attribute this significant improvement  to the use of dynamics-based invariants in the DIR stream, which provide a way of capturing the dynamics of the 3D motion from its affine projections.  Additionally, our experiments also showed that the framework works well in the related task of cross subject action recognition.  This opens up the possibility of having  widely deployable action recognition applications based on easily obtained video data, avoiding the need for special  sensors which are required to collect 3D data.



\end{document}